\documentclass[10pt,twocolumn,letterpaper]{article}

\usepackage{cvpr}
\usepackage{times}
\usepackage{epsfig}
\usepackage{graphicx}
\usepackage{amsmath}
\usepackage{amssymb}

\usepackage[font=footnotesize,labelfont=bf]{caption}

\usepackage{multirow}
\usepackage{tabularx}
\usepackage{dsfont}
\usepackage{url}
\usepackage[tight]{subfigure}
\usepackage{color}
\usepackage{subfigure}
\usepackage{amsthm}
\usepackage[ruled,vlined]{algorithm2e}
\usepackage[export]{adjustbox}

 \pdfoutput=1

\theoremstyle{definition}

\theoremstyle{remark}

\usepackage[breaklinks=true,bookmarks=false]{hyperref}

\cvprfinalcopy 

\def\cvprPaperID{****} 

\begin{document}

\title{Emergence of Selective Invariance in Hierarchical Feed Forward Networks}

\author{
  Dipan K.~Pal,~~Vishnu N. ~Boddeti,~~Marios Savvides \\
  Department of Electrical and Computer Engineering\\
  Carnegie Mellon University\\
  Pittsburgh, PA 15213 \\
  \texttt{\{dipanp, naresh, marioss\}@andrew.cmu.edu} \\
}

\maketitle

\begin{abstract}
Many theories have emerged which investigate how invariance is generated in hierarchical networks through simple schemes such as max and mean pooling. The restriction to max/mean pooling in theoretical and empirical studies has diverted attention away from a more general way of generating invariance to nuisance transformations. In this exploratory study, we study the conjecture that hierarchically building selective invariance is important for pattern recognition. We define selective invariance as carefully choosing the range of the transformation to be invariant to at each layer of a hierarchical network. For the purpose of our study, we utilize a novel method called adaptive pooling where the pooling weights are not constrained and in fact can adapt their pooling regions to the data. These networks with the adapted pooling regions maintain performances on object categorization tasks comparable to max/mean pooling networks despite being more prone to overfitting. Interestingly, adaptive pooling regions can converge to mean pooling (even when initialized with random pooling regions), find more general linear pooling schemes or even decide not to pool at all. The pooling regions that emerge from the data are not random but rather contiguous, illustrating invariance to contiguous ranges of transformations. We illustrate the general notion of selective invariance through object categorization experiments on large-scale datasets such as SVHN and ILSVRC 2012.
\end{abstract}


\section{Introduction}\label{1}

Convolutional nets (ConvNets \cite{lecun1998gradient}) have gained immense popularity over the past decade. Despite a lot of studies to improve their generalization abilities, much of their fundamental architecture remains the same. This illustrates that the basic hierarchical modules involving convolution, non-linearity and pooling operations are very effective in various domains and modalities, including vision. Nonetheless, there is much left to answer regarding what fundamental task each of these modules is achieving. In this paper, we focus on pooling, which has received relatively less attention compared to filter weights, overall architecture and training procedures.

\begin{figure}

        \centering
            \includegraphics[width=0.7\columnwidth]{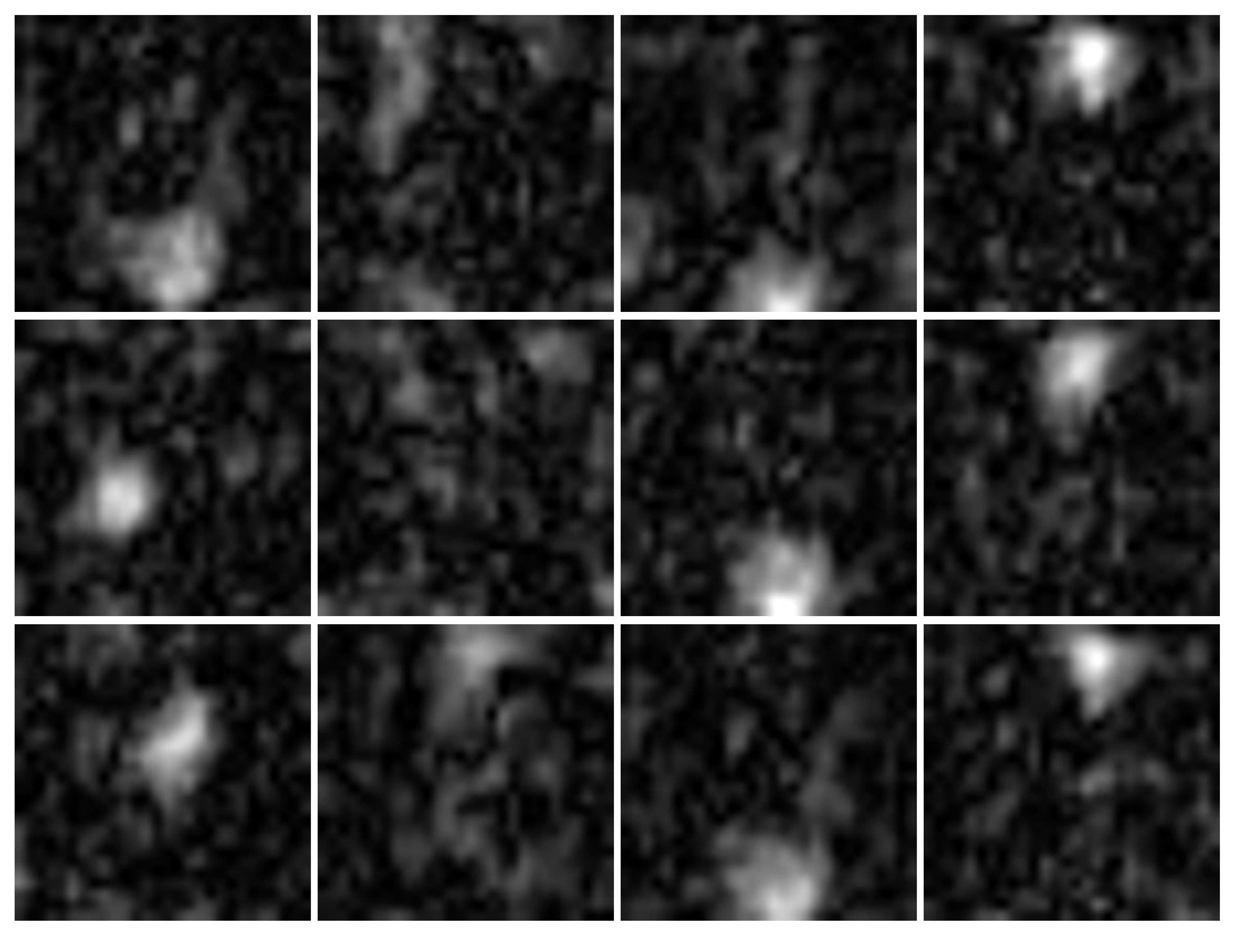}\label{fig_svhn_weight_L1}\label{fig_svhn_weights_1}

\caption{A few representative pooling weights that emerge withing pooling elements when using adaptive pooling on the SVHN dataset, in the first layer of a hierarchical feed forward network. The pooling elements with these weights are selectively invariant to the transformations within this region. Darker areas have very low weight and do not feed the input in those areas forward. The (lighter) regions also define the \textit{range} of the transformation to be invariant to. Interestingly, pooling elements converge to mean pooling despite being initialized to random pooling weights. A few prefer to be agnostic to all transformed inputs (second column). Redundancy is observed, with multiple pooling elements being invariant to similar ranges  of transformations (third column). Existence of general linear pooling schemes call for more general theories that do not specifically assume max pooling.}
\end{figure}

Traditionally, pooling is conducted over blocks or regions over the convolution map (after the convolution step) using operators such as max or mean. Essentially, pooling is part of the architecture and  defines what the ``connections" between modules are. Whereas learning the filter/kernel weights defines the ``tuning" properties in canonical architectures. Some alternative methods of pooling have recently started being explored. However, these studies approach pooling either from an engineering standpoint (making ConvNets flexible in terms of input size \cite{SpatialPyramidHeZR014}), or from a regularization perspective \cite{zeiler2013stochastic, fractionalMaxGraham14a}. Nonetheless, the efficacy of such a diverse set of approaches suggest that pooling can be effective even when implemented in various ways. Further suggesting that perhaps deeper and more fundamental objectives are at work.

One of the fundamental objective that pooling tries to optimize for, is hypothesized in many works to be generating invariance to input nuisance transformations \cite{boureau2010theoretical, saxe2011random, anselmi2015invariance}. Indeed, generating useful representations that are invariant to such transformations is arguably one of the core problems in many fields, such as vision. Pooling is thus, usually seen as a tool for introducing such invariance and it is used and implemented as such. Many successful hierarchical architectures such as ConvNets employ pooling in a deterministic and organized manner, optimized more for engineering benefits, such as fast computation and easy implementation, rather than accuracy. This is also since the specific parameters of the pooling layers (such as pooling field size) are usually selected according to heuristics and intuition due to lack of a deeper understanding of what objective pooling tries to achieve and need for specific pooling schemes. Pooling is traditionally not \textit{optimized} for the \textit{data} and only specific hyper parameters are tuned accordingly during validation to improve generalization of the network.


\section{A conjecture involving Selective Invariance to Input Transformations}

We now review and introduce some concepts useful throughout the rest of the paper. 

\textbf{Group:} A \textit{group} is a mathematical structure encoding symmetry through a set of elements along with a group operation which acts on any two elements. The structure needs to satisfy four axioms namely, closure, associativity, identity and invertibility in order for the structure to be a valid group. A group can have a finite number of elements resulting in a finite group. In cases where a group is used to model a transformation, any subset of the group can be used to define a particular \textit{range} of the transformation. 

\textbf{Invariant and Equivariant features:} Any function $f$ over $x \in \mathbb{R}^d$ is an invariant feature w.r.t a group $\mathcal{G}$ if $f(x) = f(gx)~\forall g \in \mathcal{G}$\footnote{With a slight abuse of notation, we denote by $gx$ the action of group element $g\in\mathcal{G}$ on $x$}. It is an equivariant or a covariant feature w.r.t the group if $f(x) \propto h(g)f(gx) ~\forall g \in \mathcal{G}$ where $h$ is a linear  function defined over $\mathcal{G}\mapsto \mathbb{R}$. Ideally, it is desirable for a feature to be invariant to transformations that does not change the class label of the input (intra-class transformations), but be equivariant to transformations which do (inter-class transformations).


\textbf{Invariance to transformations:} A plethora of literature exists to show that one of the core problems in pattern recognition is to generate invariance to transformations $g$ in the data, leading to significant improvements in recognition performance \cite{wood1996representation, bruna2013learning, anselmi2015invariance, leibo2014subtasks, hinton1987learning, liao2013learning}. Further, features that are invariant (even partially \emph{i.e.}invariant to a subset of $\mathcal{G}$) to the transformation group $\mathcal{G}$ allow for the sample complexity to be reduced \cite{anselmi2015invariance, liao2013learning}. Previous work such as \cite{anselmi2015invariance, leibo2014subtasks} show that even though complete invariance is unachievable in practice, (as \textit{consolation}) partial invariance holds. In this paper, we hypothesize and empirically support the claim that selective partial invariance is in fact \textit{necessary} for good recognition performance.

Indeed, the need for invariance to \textit{specific transformation ranges} in the data has been relatively understated. More formally, the specific subset  $\mathcal{G}_0$ of the group $\mathcal{G}$ to which the feature is invariant should be carefully chosen. It is more common to investigate explicit \textit{complete} invariance to transformation groups such as the rotation group and/or the translation group \cite{CohenW16, DielemanFK16}. Group invariant scattering was also proposed as a theoretical framework for modelling entire translation group invariances in ConvNets as contractions acting on the entire space globally \cite{mallat2012group}.

\textbf{Selective Invariance:} In this paper, we argue that for vision tasks (and perhaps in general), selective invariance (also equivariance) is important. Consider the range of a transformation, which refers to the extent by which the transformation is applied to a sample point. By selective invariance, we emphasize that it is necessary to be invariant to \textit{carefully chosen} parts of the range of the transformation. Additionally, it is also equally important to be equivariant to a different part of the range of the same transformation. 


To illustrate, consider the sub-task of distinguishing between 6 and 9 in an optical character recognition task. If the transformation we consider is rotation, then a $180^{\circ}$ rotation turns the 6 into a 9. Hence, the classifier not only needs to be \textit{invariant} to rotation between $-90^{\circ}$ and say  $90^{\circ}$ but, in this hypothetical situation, be \textit{equivariant} to the infinitesimally small transformation as the digit rotates just beyond the $90^{\circ}$ mark into the other class. A classifier that is completely invariant to the rotation group (\emph{i.e.} invariant to all angles from $0^{\circ}$ to $360^{\circ}$) will fail the task. This simple thought experiment illustrates the need to be invariant to parts of the range of a particular transformation while being equivariant to the other parts.

\textbf{A conjecture towards a general theory of pooling:} One of the main contributions of this paper is to provide empirical evidence and motivate a theoretical understanding of a more general form of linear pooling. We conjecture that more general forms of linear pooling exist. These more general linear pooling schemes can work comparably well to canonical pooling schemes such as max and mean pooling, thereby forcing one to not be able to ignore them while developing a theoretical model of pooling. We further conjecture as a consequence, that an additional objective for pooling operates, which suggests that though network models must build (partial) invariance hierarchically, it is sufficient to build it \textit{selectively}. This relaxes the conditions required for pooling in order for architectures to perform well, thereby allowing for more general pooling schemes. In other words, useful invariant (and equivariant) representations can be obtained by carefully choosing \textit{specific} ranges of the transformations present in the data to be invariant (and equivariant) towards. These ranges of invariance can many times be much larger and more diverse than what mean/max pooling schemes suggest and can also lead to redundancy in pooling, where multiple spatially localized pooling nodes pool across very similar ranges of transformations (see Section~\ref{sec_exp}).  Phenomenon such as these invoke the need for a more general theory of pooling.

Many previous studies have examined pooling schemes empirically and theoretically. Mean/max pooling was examined in detail in terms of discriminability in \cite{boureau2010theoretical}. The effect of max pooling on hard-vector quantized features was shown to help performance \cite{boureau2010learning}. A study more aligned towards highlighting the importance of pooling (even with random convolution filters) is \cite{jarrett2009best}. All of these efforts were however, restricted to investigating max and/or mean pooling schemes. 

\textbf{Motivating adaptive pooling:} We provide evidence for our conjecture through the use of \textit{adaptive pooling}. To this effect, we remove constraints on the canonical method of pooling. We utilize a more general linear pooling model (compared to mean pooling), and use the data itself to optimize pooling schemes. The mere existence of these more general linear pooling schemes in networks which perform comparably to max/mean pooling networks (as we find in Section~\ref{sec_exp}) show that a more general and fundamental objective for generating invariance is at play. This could suggest future theories and frameworks for ConvNets (and perhaps other deep learning algorithms) to allow for and model such general pooling schemes. Although max and mean pooling are simple to implement and work well in practice, restricting our theoretical understanding to such specialized pooling schemes might redirect attention away from more powerful and general theories for invariance and perception.


\textbf{Adaptive Pooling:} In order to investigate what kind of pooling the data requires, we propose a novel adaptive pooling layer that \textit{learns} or optimizes pooling according to the loss function and the data. The adaptive pooling layer is trained using standard back propagation along with some regularization. Our use for this layer in this paper is very specific. The adaptive pooling layer simply serves as a way to practically prove the existence of a set of network parameters (filter/kernel weights and pooling weights) that work well for the object categorization task. Selective invariance properties emerge in the pooling weights (see Fig.~2 and Fig.~6) as we find in our experiments later. We also model adaptive pooling in a group invariant framework and show how it invokes selective invariance and equivariance properties (see Section~\ref{4}).


\section{Partially Invariant Features through Partial Group Integration} \label{3}


A number of theories of invariance have emerged over the years. Most of them require some assumption regarding the structure of the transformations. One of the most common assumptions is that the transformations form a group \cite{anselmi2015invariance, laptev2015transformation, mallat2012group}. This seems valid since transformations in many fields in which the importance of invariant features seems natural such as vision, do indeed deal with common transformations that form a group, further, they are unitary (\emph{e.g.} translation and rotation). We motivate the use of adaptive pooling through such a group invariant framework. However, since in practice, all members of the group are not observed, we utilize theories which have been shown to work under partial observation of the group \footnote{A group is said to be fully observed, if during training, samples are available that have been acted upon by all members of the group. Partial observance refers to setting where only a subset of those samples are available for use.}. 

\begin{figure*}
    \begin{center}
        \subfigure[]{%
        \centering
            \includegraphics[width=0.75\columnwidth]{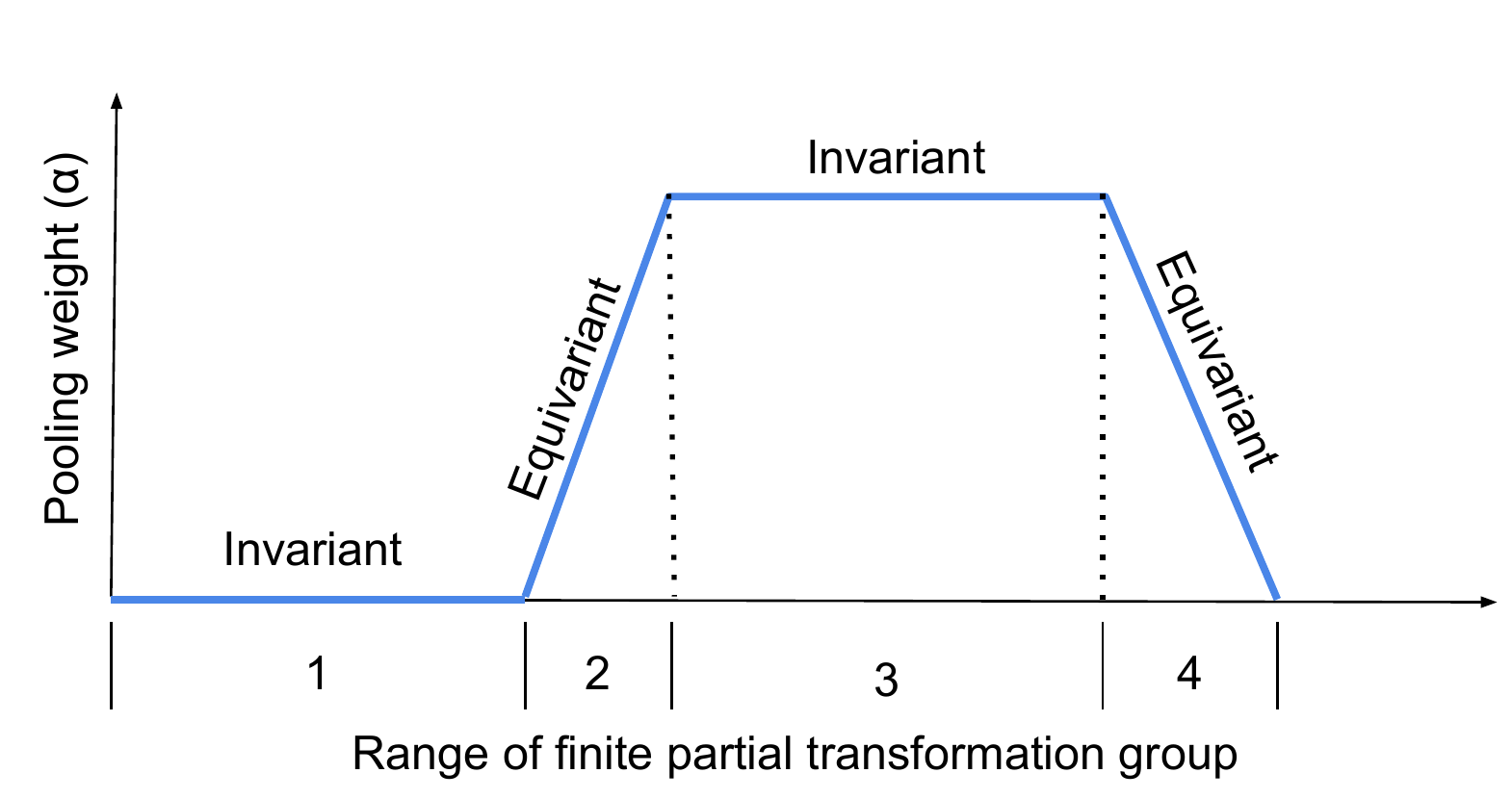}\label{fig_poolingweight}
        }%
         \subfigure[]{%
        \centering
            \includegraphics[width=0.75\columnwidth]{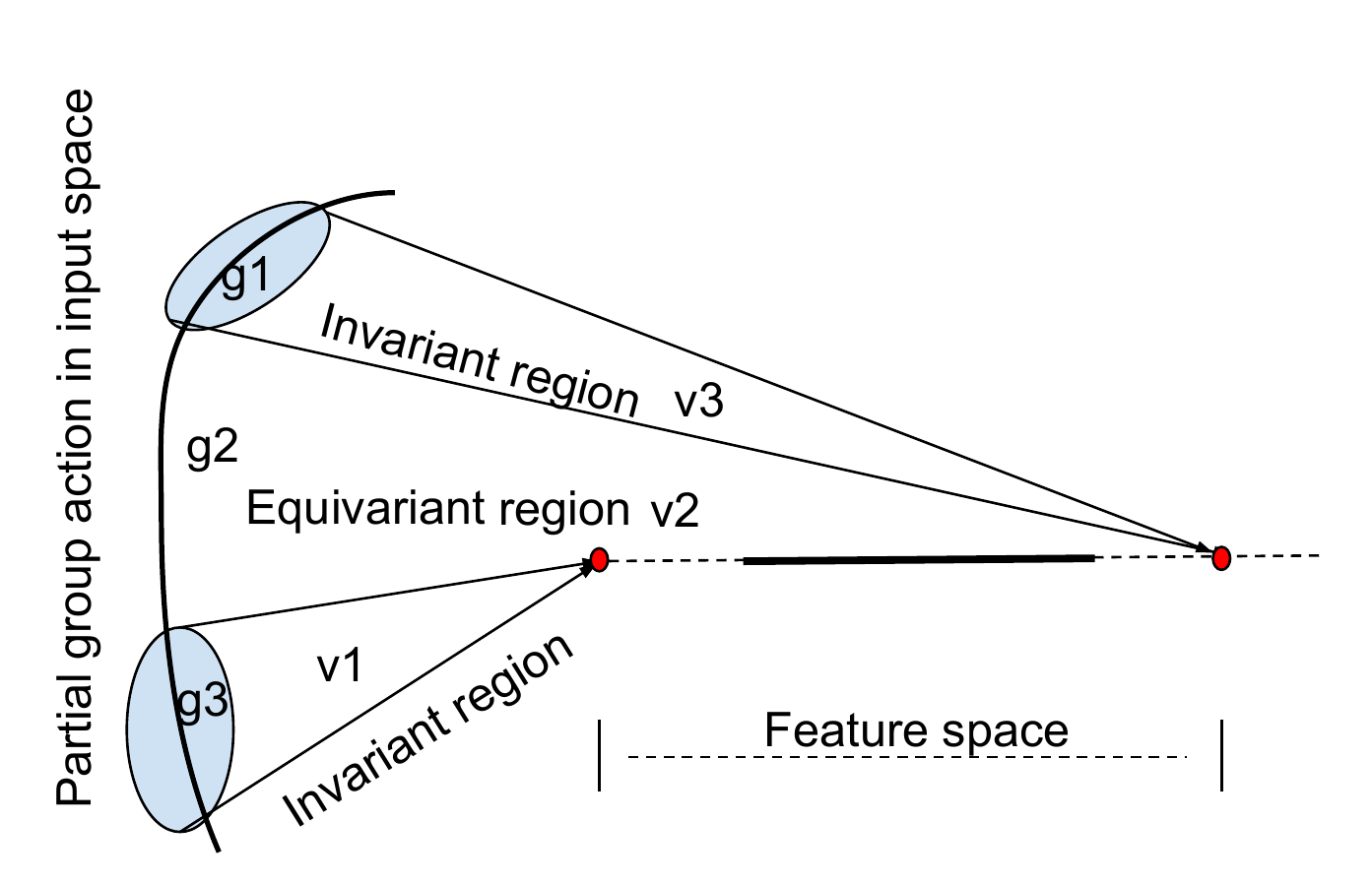}\label{fig_pooling_regions}
        }%
    \end{center}
    \vspace{-0.5cm}
\caption{(a) Illustrative depiction of pooling weights over the range of a partial transformation group. Different sections are either agnostic, invariant or equivariant depending on the weights over that particular range (subset of the partial group). (b) Diagram illustrating how subsets of the group of transformation (ranges, such as $g1, g2, g3$) map to different parts of the feature space depending on the pooling weights. $v1, v3$ depict invariant regions for a single pooling weight (corresponding to a single pooling element) which are invariant w.r.t to ranges $g1, g3$ and map them to single points (in red) for a given input. $v2$ maps $g2$ to a line (bold) since it is equivariant to $g2$.}
\end{figure*}

Consider a unitary group of transformations $\mathcal{G}$ with group elements $g$ with finite cardinality ($|G|$). 
 \textbf{Unitary Group:} A unitary group is a group with elements satisfying the unitary property, \emph{i.e.} $\langle gx, gy \rangle = \langle x, y \rangle ~\forall x, y ~\forall g\in \mathcal{G}$.
 
 We have the action of a group element $g$ on a sample point $x$ as $gx$ and following this, an orbit is generated as the set $\{ gx~|~g\in \mathcal{G} \}$. This orbit is \textit{unique} to every point since it the the set of all variations or transformations of the point as defined by  $\mathcal{G}$. A measure which introduces invariance and allows us to compare two orbits is the distribution $P_x$ induced by $\mathcal{G}$ on a sample $x$. It can be shown that  $x \sim x' \Leftrightarrow P_x = P_{x'}$
 \emph{i.e.} if two images ($x$, $x'$) are equivalent under some $g$, then their distributions are identical \cite{anselmi2015invariance}. This is important since we would like to be invariant to $\mathcal{G}$ but nonetheless have a common discriminative signature or feature for $\{ gx~|~g \in \mathcal{G}\}$. One can form such a discriminative feature by measuring properties of the distribution or trying to characterize the distribution. In order to do so, any template or filter can be utilized along with the powerful property of unitarity of the group $\mathcal{G}$.


 A single filter provides a 1-D projection of the distribution thereby providing one measurement. We can obtain many such measurements in order to be more discriminative. Such a collection of many such filters together uniquely characterizes the orbit. More importantly, unitarity of the group allows the following for a filter $t$ 
 \begin{align}
\langle gx, t \rangle = \langle  x, g^{-1}t  \rangle
\label{eq_group_property}
\end{align}
Hence, the distribution of the set $\{ \langle gx, t\rangle\},  \forall g \in G$ is exactly as that of $\{ \langle x, g^{-1}t\rangle\}, \forall g \in G$. Following this, in order to characterize the orbit of a novel sample under a group, it is not necessary to explicitly observe all its transformations under the group. Since the orbit and its corresponding distribution is invariant to $\mathcal{G}$, many possible invariant features or measures can be computed. Two sets of examples are 1) the moments of the distribution and 2) a possibly non-linear function of the inner-product (\emph{i.e.} $f(x) = \eta(\langle  x, g^{-1}t  \rangle)$, where $\eta$ is a non-linear thresholding function). Further, measures of the distribution of  $\{ \langle x, g^{-1}t\rangle\}, \forall g \in \mathcal{G}_0$ where $\mathcal{G}_0 \subseteq \mathcal{G}$ are also invariant owing to partial integrals over partial groups \cite{anselmi2015invariance}. This sets the framework for the incorporation of selective invariance. Carefully choosing $\mathcal{G}_0 $, one can control which range of the transformation a feature is partially invariant towards.

A previous study that allowed partial invariance although in a much simplistic non-hierarchical setting is  \cite{teo2007convex}. However, the incorporation of partial (non-selective) invariance was due to relaxation in an optimization framework.  It has also been argued that local features should only have enough invariance (as opposed to complete invariance) as required by the application \cite{zhang2007local}. Nonetheless, the observation was presented as a general recommendation and was not explicitly studied. 



\section{Adaptive Pooling Module for Learning Generalized Linear Pooling}\label{4}

We now describe the adaptive pooling module which generalizes mean pooling. In traditional ConvNets, mean pooling is highly structured. It is a linear operation and thus can be approximated using a matrix $A$. Each row of $A$ is called a pooling node/element with an associated pooling weight, and it performs pooling on some region of the convolution map through the inner product and hence is linear. Max pooling can be modelled in this framework by optimizing the support of the max operation instead. However, we focus on generalized linear pooling in this study.



\textbf{Notation:} We let every input to the $k^{th}$ layer be denoted by $u_k \in \mathbb{R}^m$ and the output of the layer be denoted by $v_k \in\mathbb{R}^n$. Let $\mathcal{L}$ be the loss that the network optimizes for. The adaptive pooling matrix is defined by  $A\in\mathbb{R}^{m\times n}$ with $n$ pooling elements.

We thus have $v_k = A^Tu_k$. We learn the pooling matrix using back-propagation. The gradient w.r.t to the $i^{th}$ row of $A$ (\emph{i.e.} $A_i$) then becomes.  


\begin{align}
    \frac{\partial \mathcal{L}}{\partial A_i} = \frac{\partial v_k}{\partial A_i}\left(\frac{\partial \mathcal{L}}{\partial v_k}\right) = U\alpha
\end{align}

Note that $U = \frac{\partial v_k}{\partial A_i}$ is simply a matrix with the every column as the input $u_k$. Whereas $\alpha = \frac{\partial \mathcal{L}}{\partial v_k}$ is a vector of coefficients. Now recall that the response map $u_k$ (\emph{i.e.} input to the adaptive pooling layer) in the case of ConvNets is a collection of the inner products of the input map of the previous convolution layer with a convolutional kernel $\omega$ passed through a non-linearity $\eta$. This can be modelled as a partial translation group $\mathcal{G}_T$ (composed of $m$ translation operators) that acts upon a convolutional filter $\omega$ to form the set $\{ g_T\omega ~|~ g_T\in \mathcal{G}_T \}$. A translation group is a group composed of translation operators.  The elements of $u_k$ are then in the form of the set $\{ \eta (\langle g_T\omega, x \rangle) ~|~ g_T\in \mathcal{G}_T \}$. Here, $x$ is the input to the previous convolution layer. Other network architectures which incorporate more transformations (such as \cite{gens2014deep, dieleman2016exploiting}) can also be modelled in this framework as long as the transformation is unitary.



The response of the adaptive pooling layer, \emph{i.e.} the $i^{th}$ pooling element at the $k^{th}$ layer with a pooling weight vector $\alpha^i$ computes 
\begin{align}
    v_{ki} = \sum_{j=1}^M \alpha^i_{j} \eta (\langle g_{Tj}\omega, x \rangle) = \sum_{j=1}^M \alpha^i_j \eta (\langle \omega, g^{-1}_{Tj} x \rangle) ~~~\forall g_{Tj}\in \mathcal{G}_T\label{eq_inv}
\end{align}
The second equality holds from the unitary property of $\mathcal{G}$ and the fact that $\mathcal{G}$ is a group (partial groups have corresponding inverses which also form a partial group). Hence, the pooling vector $\alpha$  effectively pools over the transformations of the input. This computes a measure of the orbit of the input under the partial group. In the case when $\alpha$ is a vector with the same coefficients, this is exactly the group integral over the finite partial translation group $\mathcal{G}^{-1}_T$. It has been shown that a non-linear feature of the form as Equation~\ref{eq_inv} is partially invariant given a finite partial group \cite{anselmi2015invariance}. In practice however, $\alpha$ also includes significantly varying coefficients, thus introducing selective invariance (and equivaraince) to certain transformation ranges in the partial group. 
\begin{figure*}
    \begin{center}
        \subfigure[]{%
        \centering
            \includegraphics[width=0.7\columnwidth]{figs/SVHN_weights_layer_1.pdf}\label{fig_svhn_weight_L1}\label{fig_svhn_weights_1}
        }%
         \subfigure[]{%
        \centering
            \includegraphics[width=0.7\columnwidth]{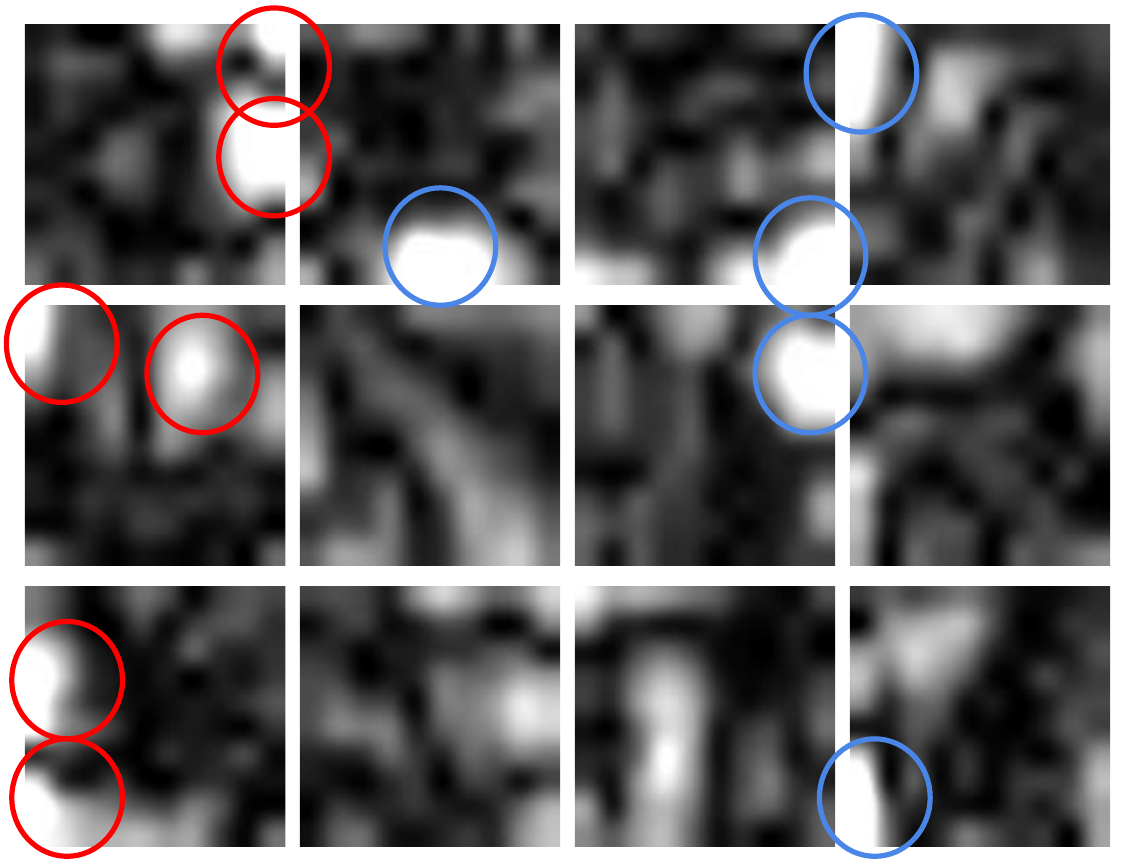}\label{fig_svhn_weight_L2}\label{fig_svhn_weights_2}
        }%
    \end{center}
    \vspace{-0.5cm}
\caption{A few representative pooling weights learned using adaptive pooling on the SVHN dataset. (a) Pooling weights from layer 1. Interestingly, pooling elements converge to mean pooling despite being initialized to random pooling weights. A few prefer to be agnostic to all transformed inputs (second column). Redundancy is observed, with multiple pooling elements being invariant to similar ranges  of transformations (third column). (b) Pooling weights from layer 2. More interesting selective invariance to specific transformations emerge. Many pooling elements are selective to large contiguous ranges of transformations (circled in blue) whereas a few elements prefer to be invariant to multiple contiguous ranges (multiple pooling regions, first column, circled in red).}
\end{figure*}
In order to reduce overfitting and improve learnability of the pooling weights, we constrain the $\ell_1$ norm of each pooling element to be 1\footnote{Recall that the goal is to find pooling architectures and parameters that work well in practice which deviate from the standard pooling schemes. The regularization is thus justified, since  merely proving the existence of such parameterizations is enough to showcase a more general form of pooling and hint towards the more fundamental goal of selective invariance.}. Under such regularization, $\alpha$ will compose of non-zero elements as well as elements that are negligibly small. This defines the support of the range of the transformations that the pooling element is invariant to. Fig.~\ref{fig_poolingweight} illustrates the range of transformations that the pooling weight is invariant and equivariant to for a typical pooling response.  Part 1 of the pooling element in Fig.~\ref{fig_poolingweight} contains near zero weights, thus the pooling element is agnostic to all input in that range of transformation. Part 3 is comprised of a range which has approximately constant weight. This part will be invariant to that particular range of the transformation while providing an invariant descriptor of the orbit of the input under the partial group corresponding to that range. 


Part 2 and 4 in Fig.~\ref{fig_poolingweight} however, are parts which are not invariant, but rather approximately equivariant or covariant wherein the response of the pooling element is linearly proportional to the group element \emph{i.e.} $v_{ki} \propto h(g_{i}) \eta (\langle g_{i}\omega, x \rangle)~\forall g_{i}\in \mathcal{G}^{-1}_{\text{equi}T} \subseteq \mathcal{G}^{-1}_T$, where $\mathcal{G}^{-1}_{\text{equi}T}$ defines the range of transformations in $\mathcal{G}^{-1}_T$ that the pooling weight $v_{ki}$ is equivariant to, and $h$ is a linear function defined over $\mathcal{G} \mapsto \mathbb{R}$. Such a linear function $h$ exists since the group $\mathcal{G}_T$ is composed of transformations that are linearly related to each other. The selectivity in each pooling element is also derived from its equivariant or covariant responses to certain ranges of the transformation apart from the invariant responses to other ranges. 


\textbf{Potentially redundant pooling elements:} Adaptive pooling is a generalization of mean pooling. Unlike mean pooling however, every pooled element is not restricted to pool over a pre-defined $p\times p$ grid (usually with $p=2, 3$). Each pooling element can in fact, in theory, pool over the entire response map of the previous layer. The overlap between different pooling elements is not pre-defined to be complimentary and can in fact be similar for multiple pooling elements, thereby introducing redundancy. This is a phenomenon we do indeed observe in our experiments (see Section~\ref{sec_exp}). This is somewhat counter-intuitive to what we might expect to provide a good representation, wherein each element might be expected to capture a different feature or aspect of the data. Nonetheless, in practice such a configuration emerges to perform just as well, and hence it provides more insight into pooling and representation. 

\textbf{Need for redundancy:} Redundancy in pooling elements could be hypothesized to provide and retain significant activations at any given layer, where pooling reduces the size of the activation map up the hierarchy. Hence pooling elements must compete and/or cooperate to utilize the limited space available in the activation map for activations more useful for the task. Having redundant pooling elements that are located in the near vicinity of each other spatially preserve the locality and contiguity of the activation of the object.



\section{Emergence of Selective Invariance and Redundant Pooling}\label{sec_exp}

We use standard ConveNets architectures while replacing the mean and max pooling layers with adaptive pooling. We train the networks to minimize the logistic soft max loss on large-scale classification benchmarks such as the Street View House Numbers (SVHN) and the ImageNet Large Scale Visual Recognition Challenge (ILSVRC) 2012 datasets. Mean and max pooling field sizes were fixed at $2\times 2$ for all experiments.


\begin{figure}\label{fig_svhn_samples}
\centering
\includegraphics[width=0.4\columnwidth]{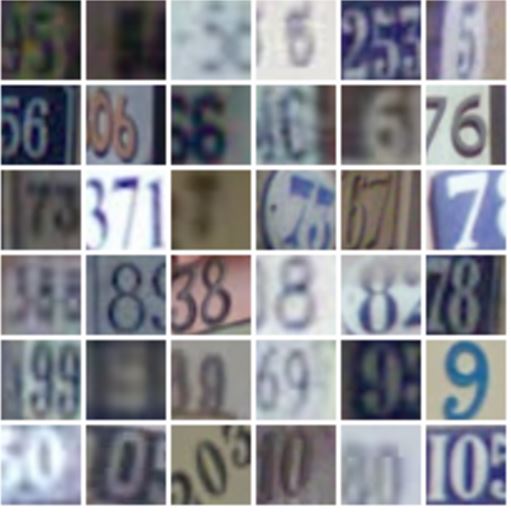} 
    \vspace{-0.2cm}
\caption{Representative samples from the SVHN dataset.}
\end{figure}

\subsection{Street View House Numbers (SVHN)}
 The SVHN dataset has 10 classes corresponding to 10 digits and a training and testing data size of about 73,000 and 26,000 samples. For this dataset we use a network with two convolution layers (64 filters of 5$\times$5 each ) each followed by an adaptive pooling layer. The last two layers were fully connected (1600 and 128 nodes). Non-linear layers were used after every convolution. The network parameters were randomly initialised including the adaptive pooling weights. All networks were trained using dropout for 400 epochs.

\begin{figure}
     \begin{center}
     \begin{tabular}{c}
        \subfigure[]{%
        \centering
            \includegraphics[width=0.7\columnwidth]{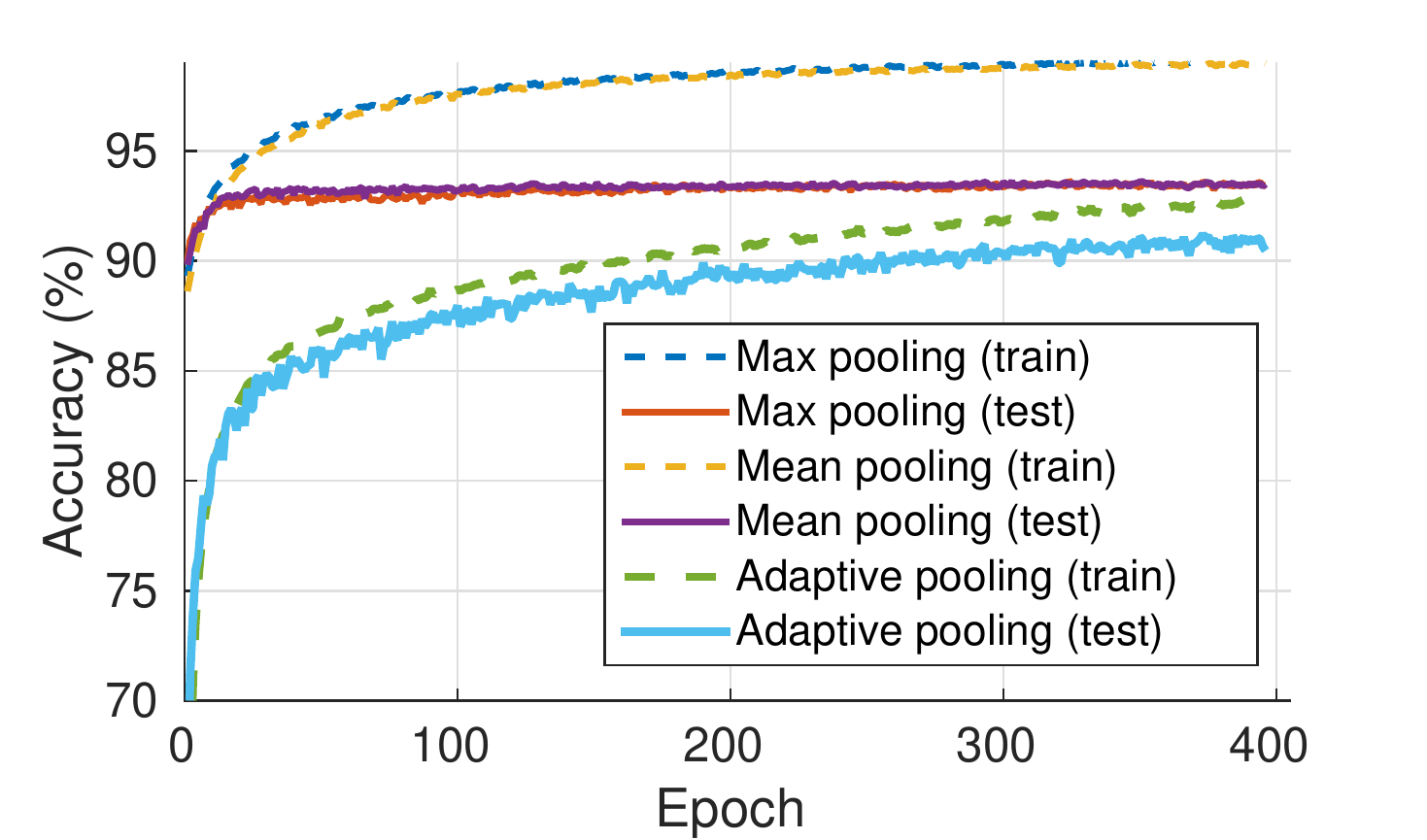}\label{fig_svhn}
        }\\%
         \subfigure[]{%
        \centering
            \includegraphics[width=0.75\columnwidth]{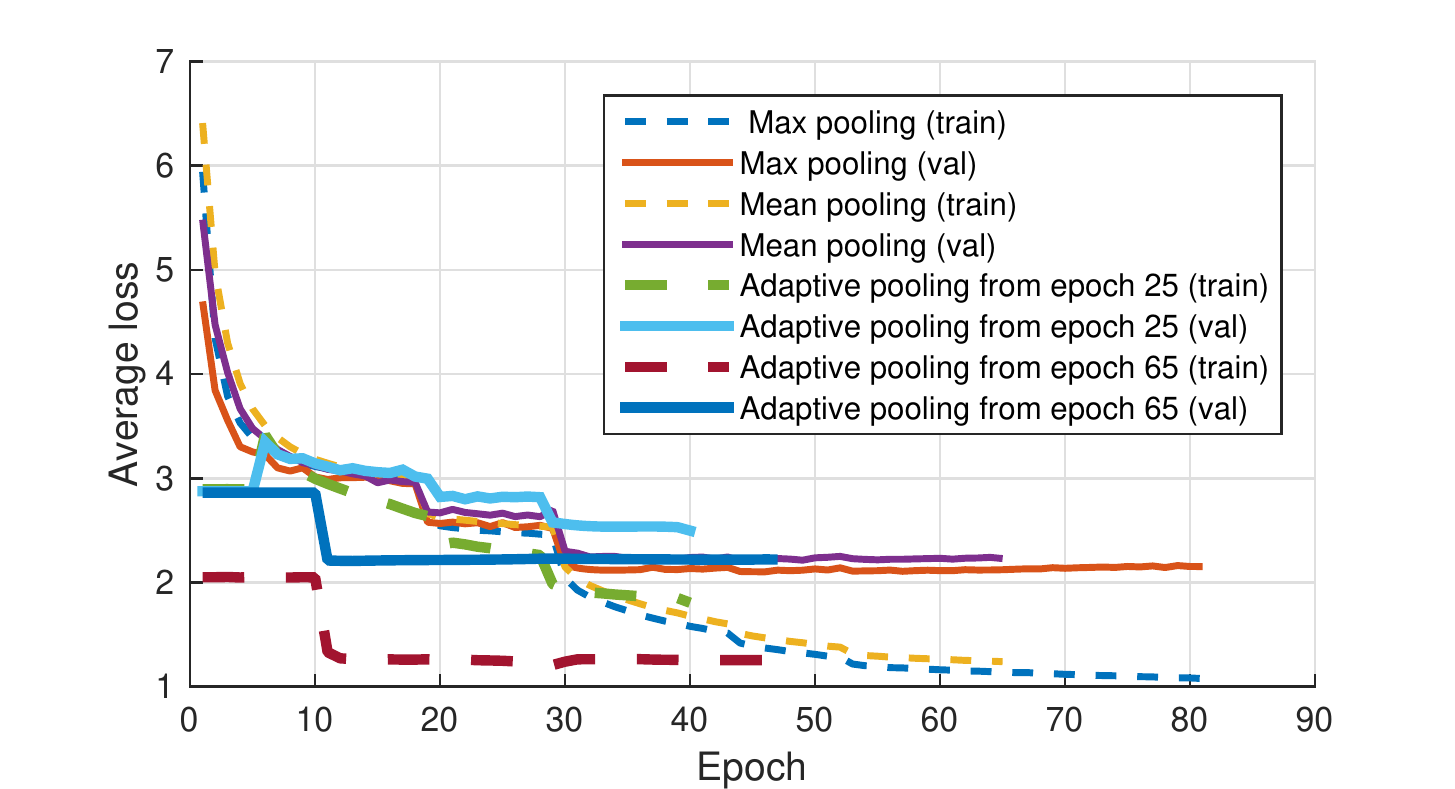}\label{fig_imagenet}
        }%
        
        \end{tabular}
    \end{center}
    \vspace{-0.5cm}
\caption{(a) Train and test accuracies (\%) on the SVHN test data for ConvNet architectures utilizing max, mean and adaptive pooling (initialized with random pooling weights) (b) Average loss on train and validation data in the ILSVRC 2012 dataset for max, mean and adaptive pooling. Adaptive pooling was initialized with mean pooling weights and convolutional weights from models at the  $25^{th}$ and $65^{th}$ epoch from the mean pooling run.}\label{fig_svhn_results}
\end{figure}

 \textbf{Results:} Fig.~\ref{fig_svhn_results} shows the progression of train and test accuracies for all three pooling schemes. Although adaptive pooling suffers in performance initially, it recovers over epochs achieving close to $\sim 91\%$ accuracy compared to the mean/max pooling result of $\sim 93\%$. The adaptive pooling network, in this particular experiment, was initialized using random weights. One might expect the network to have difficulty learning given the large number of parameters, however, given the easier task (compared to an even larger scale classification task such as ILSVRC 2012), the network gradients are informative and the network performance improves. 
 

 Fig.~\ref{fig_svhn_weights_1} and Fig.~\ref{fig_svhn_weights_2} shows some representative pooling weights from the final model learned using adaptive pooling. We find that at layer 2, pooling elements significantly deviate from the canonical mean pooling scheme. More interestingly, mean pooling emerges in the layer 1 pooling weights despite the random pooling weights initialization. The pooling elements adapt to the transformations present in the dataset by being selectively invariant to multiple ranges of transformations (first column of Fig.~\ref{fig_svhn_weights_2} circled in red) or larger contiguous ranges (circled in blue). Also interestingly, a few pooling elements in layer 1 tune to being completely agnostic to all inputs (second column in Fig.~\ref{fig_svhn_weights_1}). This seems to be an artifact of the dataset which is composed of optical characters that usually lie near the center as shown in Fig.~3. The backgrounds around the digits near the edges are irrelevant to the task and hence transformations of those areas are not useful. The adaptive pooling elements learns to be completely agnostic to transformations of all inputs in that locality.

 \textbf{Comparison of Adaptive Pooling to max/mean pooling performance.} Adaptive pooling has many more parameters than max/mean pooling making it susceptible to over-fitting and also having the effects of local minima be more pronounced. There exist methods to help adaptive pooling achieve better performance through regularization etc. However this is not the goal of this study. The goal of the study is to let patterns emerge from the data with minimal regularization and to use very simple and canonical optimization techniques such as gradient descent.

\subsection{ ImageNet Large Scale Visual Recognition Challenge (ILSVRC) 2012}

 \begin{figure}\label{fig_imagenet_layer_21}
\centering
\includegraphics[width=0.5\columnwidth]{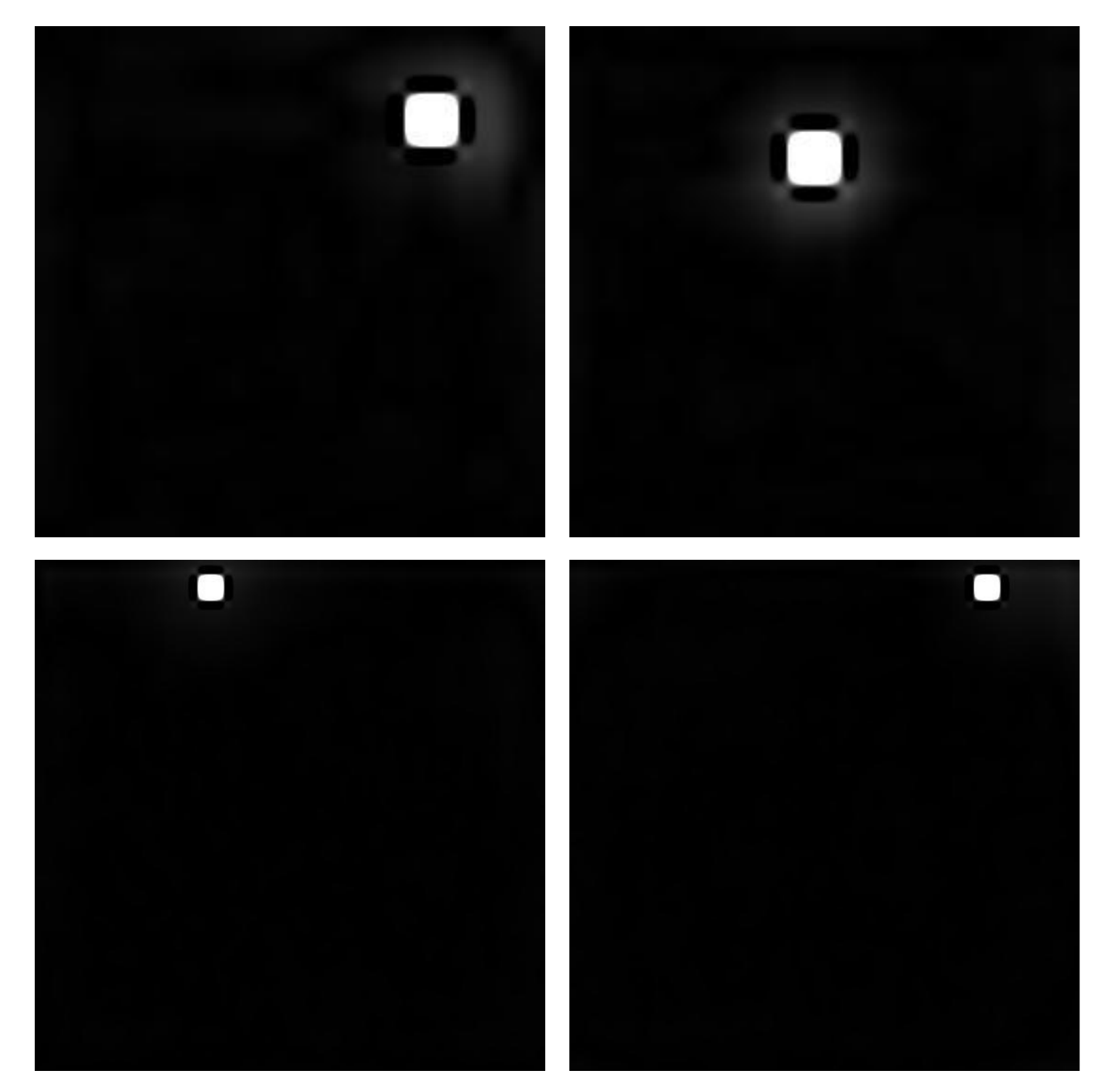} 
    \vspace{-0.2cm}
\caption{Representative pooling weights learnt using adaptive pooling from layers 1 (bottom row) and 2 (top row) of the adaptive pooling enabled AlexNet on ILSVRC 2012. Layers 1 and 2 preserved mean pooling.}
\end{figure}
The ILSVRC 2012 challenge has about 1,000 classes and over 1.2 million images for training and about 50,000 images for validation. We use the standard AlexNet architecture \cite{krizhevsky2012imagenet} for this task. We benchmark against standard mean and max pooling. To incorporate adaptive pooling, we replace all three pooling layers in AlexNet with adaptive pooling.

\textbf{Initialization:}  Convolution filters for baseline networks with mean and max pooling are always initialized randomly. We initialized the network with adaptive pooling in two ways. First, we tried initializing adaptive pooling parameters along with convolution parameters randomly. This resulted in extremely slow learning owing to the increased number of parameters, no regularization and a harder classification task. Second, we pre-trained AlexNet with mean pooling for 25 and then 65 epochs (with convolutional layers initialized randomly) and then replaced the mean pooling layers with adaptive pooling for the models at epoch 25 and epoch 65. The adaptive pooling layers were then initialized with mean pooled weights and training continued.

\begin{figure*}
\begin{center}
    \begin{tabular}{c c c c}

        \subfigure[]{%
            \includegraphics[width=0.45\columnwidth]{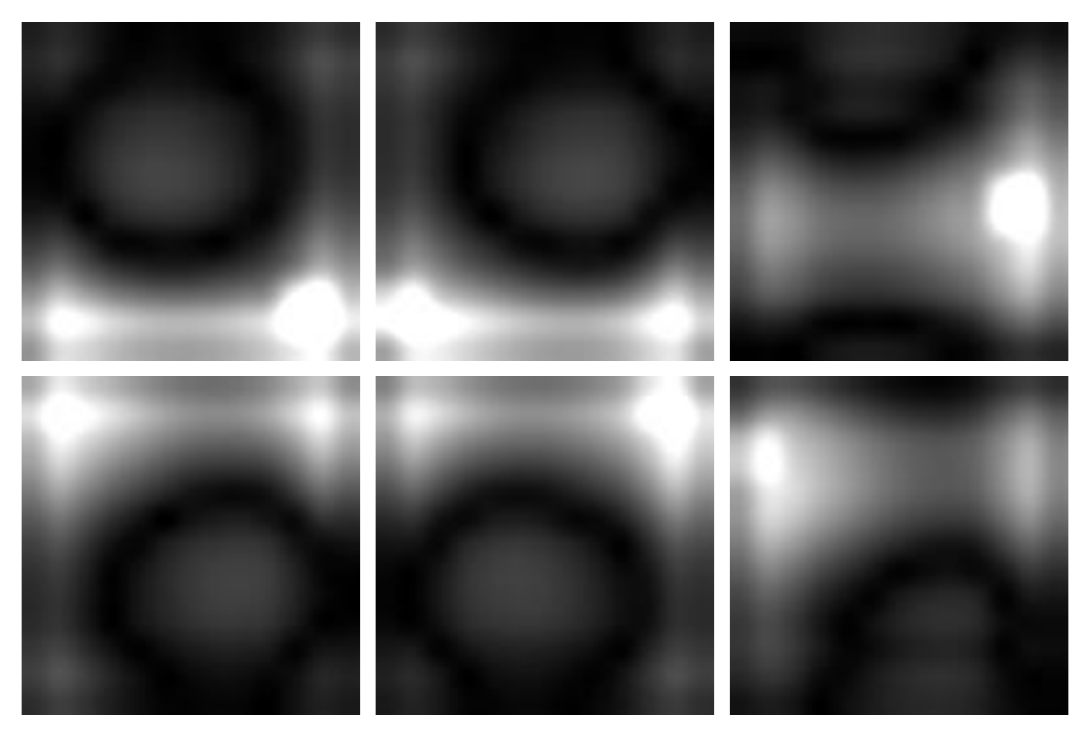}\label{fig_imagenet_weights_1}
        } &
         \subfigure[]{%
            \includegraphics[width=0.45\columnwidth]{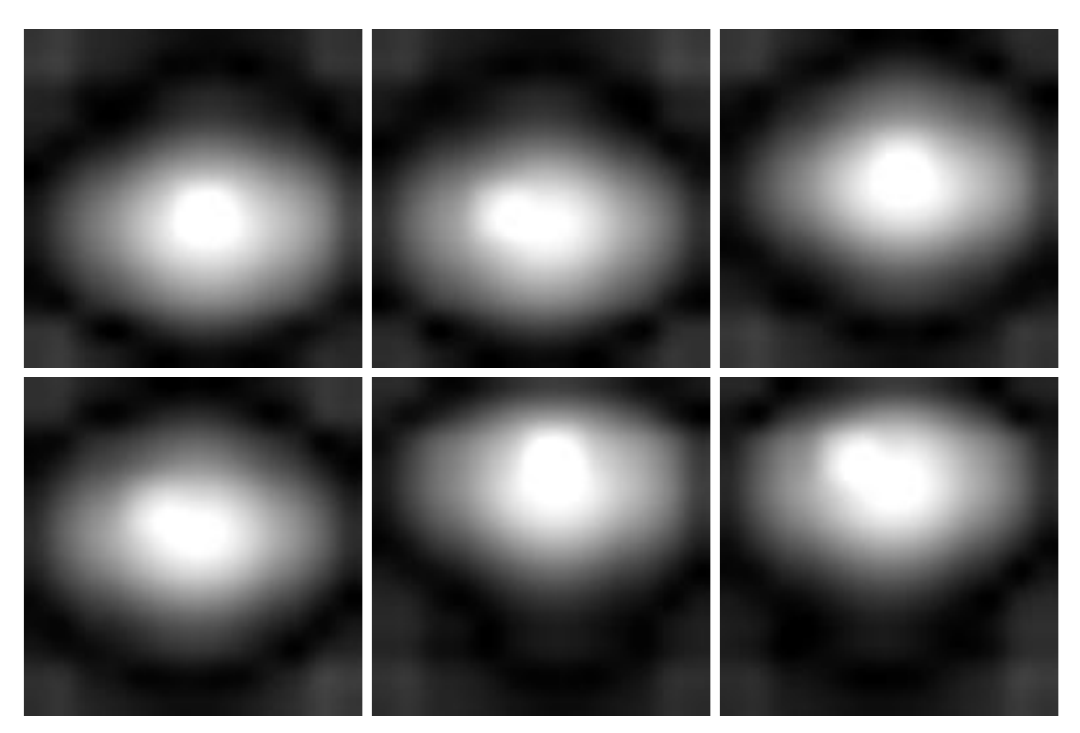}\label{fig_imagenet_weights_2}
        }&
        
        \subfigure[]{
            \includegraphics[width=0.45\columnwidth]{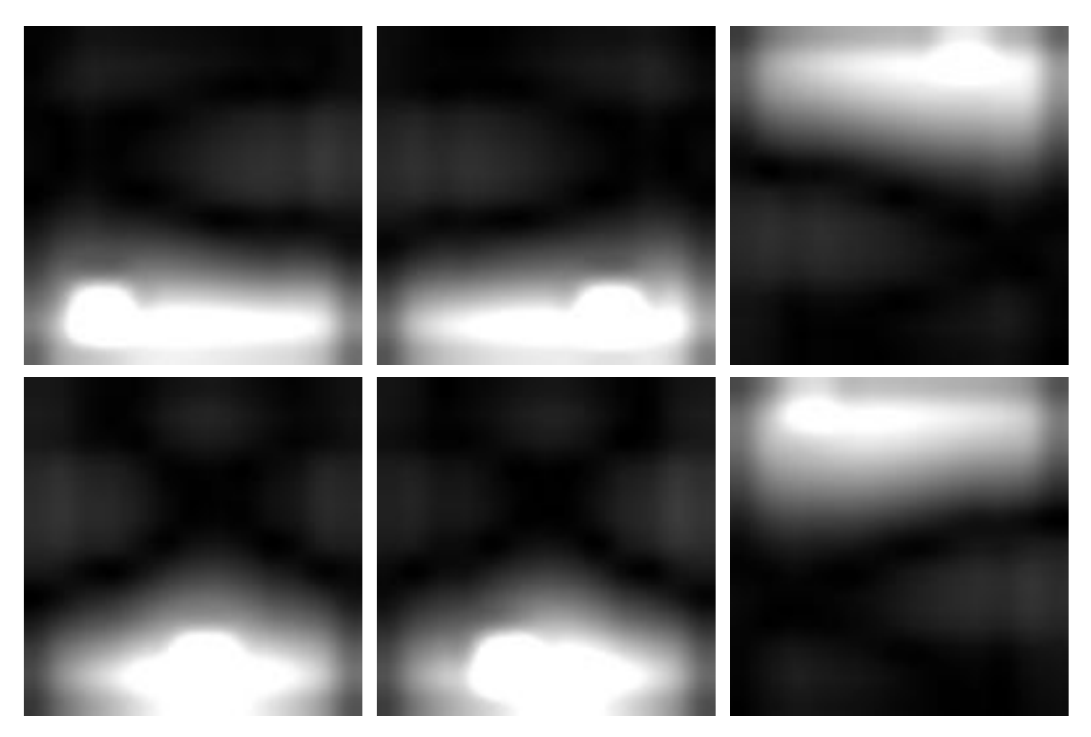}\label{fig_imagenet_weights_3}
        }&
         \subfigure[]{%
            \includegraphics[width=0.45\columnwidth]{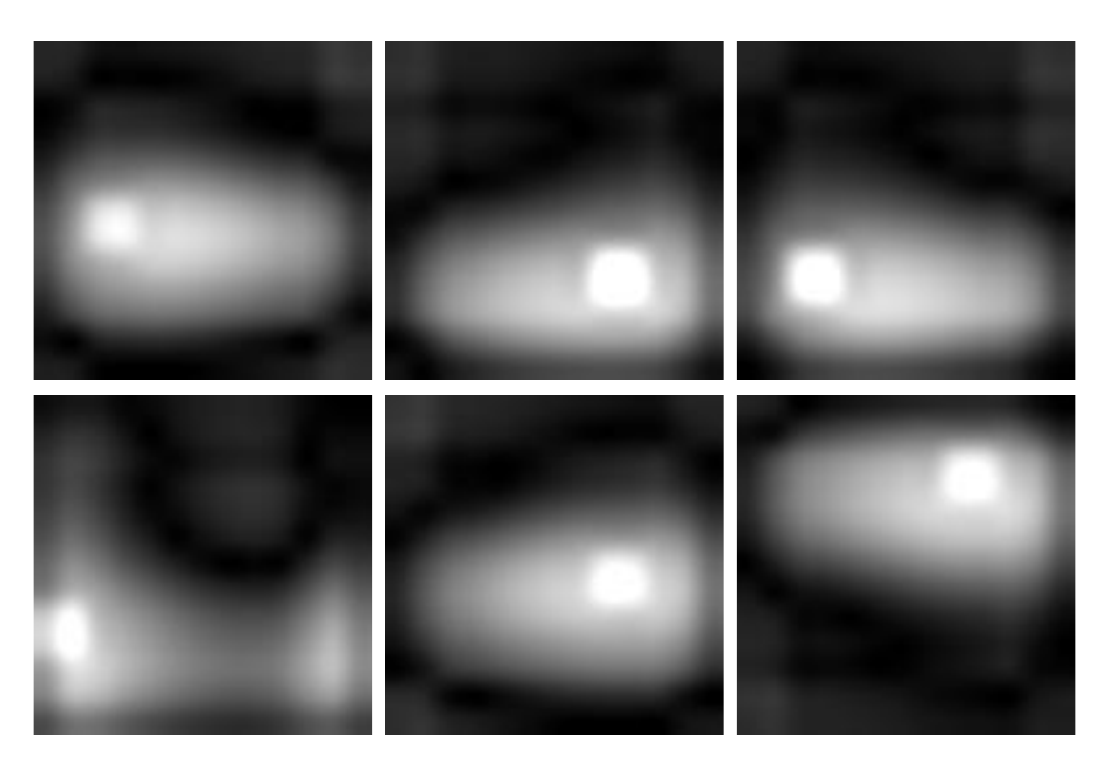}\label{fig_imagenet_weights_4}
        }%
    \end{tabular}
    \end{center}
    \vspace{-0.5cm}
\caption{Representative pooling weights from layer 3 of the adaptive pooling enabled AlexNet on ILSVRC 2012. Interesting kinds of selective invariance to transformations emerge. (a) Selective invariance to multiple disjoint ranges of transformations (pooling on spatially discontinuous regions ) (b) Selective invariance to a single large range (pooling over a large spatially contiguous region) with multiple elements that are located close by. They are invariant to the same range (redundant pooling over large ranges) (c) Multiple elements that are highly selectively invariant to specific ranges (redundant pooling over specific ranges) (d) Highly selectively pooling  elements (essentially mean pooling was preserved). }
\end{figure*}

\textbf{Results:} Fig.~\ref{fig_imagenet} shows the average training and validation loss on ILSVRC 2012 for max, mean and adaptive pooling. After the learning rate drop during training (after first 10 epochs), adaptive pooling (epoch 65) almost matches mean and max pooling despite the large increase in the number of parameters. Initialization of the network to mean pooling weights and the convolution layers to pre-trained convolutional filters  (from epoch 25 and 65 respectively) help overcome adverse effects that accompany the increase. Fig.~2 and Figs.~\ref{fig_imagenet_weights_1}, \ref{fig_imagenet_weights_2}, \ref{fig_imagenet_weights_3}, \ref{fig_imagenet_weights_4} show some representative pooling weights from the adaptive pooling layers (layers 1, 2 and layer 3 respectively) of the model learned using AlexNet pre-trained up until epoch 65. The adaptive pooling layer was fine tuned for about 50 additional epochs.

\section{Discussion}


\textbf{Emergence of Selective Invariance:} Our first observation through the SVHN (see Fig.~\ref{fig_svhn}) and the ILSVRC 2012 (see Fig.~\ref{fig_imagenet}) experiments,  is that adaptive pooling can perform comparably to max/mean pooling schemes. This validates the efficacy of the generalized linear pooling parameterization that adaptive pooling finds. In many cases, the pooling weights were found to deviate significantly from mean pooling schemes. Additionally, a few pooling elements were found to become completely agnostic to all inputs despite being initialized to random pooling weights (see Fig.~2). Both these observations illustrate the emergence of selectively invariant pooling elements in the networks.

\textbf{Pooling elements at lower layers  tend to be selectively invariant to smaller contiguous ranges of transformations:}  It is interesting to find that adaptive pooling elements initialized with random pooling weights converge to mean pooling at layer 1 for SVHN (see Fig.~\ref{fig_svhn_weights_1}). Further, even though adaptive pooling for AlexNet on ILSVRC 2012 was initialized to mean pooling weights, mean pooling was preserved for layers 1 and 2 (see Fig.~5). This leads to an observation that invariance should be generated locally for local features in low level representations. This agrees with the hypothesis that low level object parts and their features have fewer transformations that they can undergo, which have a smaller support over the input space and hence are local. Pooling elements invariant to those transformations are also localized. Mean pooling therefore seems to be a good approximation for invariant features at lower levels in hierarchical networks.


\textbf{Pooling elements at higher layers tend to be selectively invariant to larger (possibly non-contiguous) ranges of transformations: } As a general trend we also observe that the pooling elements at higher layers such as layer 3 for AlexNet (see Fig.~6) and layer 2 for the SVHN network (see Fig.~\ref{fig_svhn_weights_2}) need specialized invariant features since the pooling weights deviate significantly from mean pooling. This is despite AlexNet pooling layers being initialized to mean pooling weights. This also agrees with the hypothesis that high level object parts and complete objects can undergo a more complex set of transformations. Pooling elements that are selectively invariant at higher layers can sometimes be redundant and be invariant to extremely large contiguous ranges or even multiple smaller ranges. Hence, pooling at higher layers needs more careful handling.  Perhaps mean pooling at higher layers is sub-optimal and more effective pooling strategies that are selectively invariant could help improve performance of these networks in general.

{\small
\bibliographystyle{ieee}
\bibliography{adaptivepooling}
}

\end{document}